\documentclass[letterpaper]{article} 
\usepackage{aaai23-r2hcai}  
\usepackage{times}  
\usepackage{helvet}  
\usepackage{courier}  
\usepackage[hyphens]{url}  
\usepackage{graphicx} 

\usepackage{natbib}  
\usepackage{caption} 
\usepackage{algorithm}
\usepackage{algorithmic}

\usepackage{newfloat}
\usepackage{listings}

\DeclareCaptionStyle{ruled}{labelfont=normalfont,labelsep=colon,strut=off} 
\floatstyle{ruled}
\newfloat{listing}{tb}{lst}{}
\floatname{listing}{Listing}

\usepackage{fancyhdr}
\fancypagestyle{firstpagehf}
{
   \fancyhf{}
   \fancyhead[HC]{AI for Bangla 2.0}
   \fancyfoot[C]{\thepage}
}

\title{Connecting the Dots: Leveraging Spatio-Temporal Graph Neural Networks for Accurate Bangla Sign Language Recognition}

\author{Haz Sameen Shahgir$^{1}$, Khondker Salman Sayeed$^{1}$,\\ Md Toki Tahmid$^{1}$, Tanjeem Azwad Zaman$^{1}$\\ Md. Zarif Ul Alam$^{1}$\footnote{Corresponding Author.\\  E-mail: 1705010@ugrad.cse.buet.ac.bd}
}
\affiliations {
    \textsuperscript{\rm 1} Department of CSE, BUET \quad
}

\usepackage{bibentry}

\begin{document}
\thispagestyle{firstpagehf}
\maketitle

\begin{abstract}

Recent advances in Deep Learning and Computer Vision have been successfully leveraged to serve marginalized communities in various contexts. One such area is Sign Language - a primary means of communication for the deaf community. However, so far, the bulk of research efforts and investments have gone into American Sign Language, and research activity into low-resource sign languages - especially Bangla Sign Language - has lagged significantly. In this research paper, we present a new word-level Bangla Sign Language dataset - BdSL40 - consisting of 611 videos over 40 words, along with two different approaches: one with a 3D Convolutional Neural Network model and another with a novel Graph Neural Network approach for the classification of BdSL40 dataset. This is the first study on word-level BdSL recognition, and the dataset was transcribed from Indian Sign Language (ISL) using the Bangla Sign Language Dictionary (1997). The proposed GNN model achieved an F1 score of 89\%. The study highlights the significant lexical and semantic similarity between BdSL, West Bengal Sign Language, and ISL, and the lack of word-level datasets for BdSL in the literature. The dataset and source code are publicly available.

\end{abstract}

\section{Introduction}

Sign language is a vital mode of communication for the deaf community. While sign language dictionaries provide a foundation for learning sign language, the practical application of sign language varies in terms of phonological, morphological, grammatical, and lexical aspects \footnote{https://github.com/Patchwork53/BdSL40\_Dataset\_AI\_for\_Bangla\_2.0\_Honorable\_Mention}. The diversity of sign languages is influenced by regional expressions and language alphabets, resulting in multiple sign languages, such as American, Arabic, French, Spanish, Chinese, and Indian. In Bangladesh, the lack of proper devices or methods that can serve as interpreters makes it necessary for hearing individuals to learn sign language to communicate with the deaf. However, the complexity of the Bangla Sign language, which utilizes both hand and body gestures, poses a challenge for individuals seeking to learn it. Consequently, deaf individuals have difficulty teaching their sign language to hearing individuals, creating a distance between the deaf and the broader society. Therefore, there is a pressing need to develop an interpreter that can translate sign languages into text or speech, enabling deaf individuals to communicate effectively with society. This necessity has made the recognition of Bangladeshi sign language (BdSL) a significant and challenging topic in the field of computer vision and machine learning.

While current works on sign language recognition are classified into two categories - isolated and continuous sign language recognition, the lack of available datasets has hindered research into Bangla Sign Language (BdSL) recognition. In particular, there has been no concerted effort to collect word-level data on BdSL, which has limited the application of machine learning in this area.

Our contributions are summarized as follows:

\begin{itemize}

\item We present the first word-level Bangla Sign Language dataset (BdSL40), consisting of 611 videos over 40 BdSL words, with 8 to 22 video clips per word. This dataset addresses the lack of available word-level datasets for BdSL, enabling research into sign language recognition using machine learning.

\item We propose a model for the classification of BdSL40 using a 3D Convolutional Neural Network (3D-CNN) \cite{tran2015learning}, which achieved a peak accuracy of 82.43\% on an 80-20 split of the dataset. This provides a strong baseline for future research in this area.

\item We also propose a novel method for BdSL40 classification by extracting key points from the videos and constructing a spatiotemporal graph. We then use a Graph Neural Network \cite{yu2017spatio} to classify the dataset, achieving a peak accuracy of 89\% on an 80-20 split of the dataset. This method provides an alternative approach to the classification of sign language recognition and demonstrates the potential of GNNs in this area.

\end{itemize}

\newpage

\section{Dataset Creation}

Owing to geographical proximity and significant cultural crossover, the sign languages of Bangladesh (BdSL), West Bengal (WBSL), and India (ISL) bear striking similarities with one another. BdSL and WBSL have high semantic similarity and are mutually comprehensible while BdSL and ISL have 75\% lexical similarity but differ in their meaning \citep{johnson2016distinction}. For example, the sign for ``Fish" in ISL is the same as for ``Tortoise" in BdSL. Out of the three mentioned sign languages, only ISL has a sizable word-level dataset available to the public: INCLUDE by \citet{sridhar2020include}. We went over the INCLUDE dataset and consulted the Bangladesh Sign Language Dictionary (1997) to find the Bangla meaning of the signs in the INCLUDE dataset. From the 263 words in INCLUDE, we were able to collect 40 terms of which 28 words had the same meaning and sign in both ISL and BdSL, and 12 had the same sign but a different meaning in BdSL. It is to be noted that the 40 words had the same signing motions in both BdSL and ISL. We noticed similarities between other words but did not include them in BdSL40 since the signing motions were not the same.

In total, there are 611 videos over 40 BdSL words, with 8 to 22 video clips per word.

\section{Proposed Methodology}

We present the methodology for our 2 different approaches in this section.

\subsection{Video ResNet}

In this section, we describe our methodology for classifying the Bangla Sign Language 40 (BdSL40) dataset using Video ResNet \cite{tran2018closer}. The proposed methodology consists of two main stages: dataset preprocessing and model training.

\paragraph{Dataset Preprocessing.}

The BdSL40 dataset comprises 611 videos. Each video has a resolution of 1080x1920 and a frame rate of 30 frames per second. Before training the model, we first preprocess the dataset to ensure that the data is in a format suitable for training. First, we split the dataset into training and testing sets with an 80-20 ratio. Then, we extracted the frames from each video and resized them to 100x100 to reduce the dimensionality of the data.

\paragraph{Model Training.}

We trained the model using the Video ResNet architecture, which is a variant of the ResNet architecture that is specifically designed for video data. The Video ResNet architecture consists of three main components: a 3D convolutional backbone, a temporal pooling layer, and a fully connected layer. The 3D convolutional backbone extracts spatio-temporal features from the input video frames. The temporal pooling layer aggregates the spatio-temporal features across time to produce a fixed-length feature vector for each video clip. The fully connected layer then maps the feature vector to the output classes.

During training, the first 6 and last 8 frames of each video were skipped to remove any redundant information. Frames were normalized by subtracting the mean value of 0.5 and dividing by the standard deviation of 0.5 for each color channel. The model was trained for 120 epochs with a batch size of 64 and a learning rate of 5e-5. 

\begin{figure*}
  \centering
  \includegraphics[width=\linewidth]{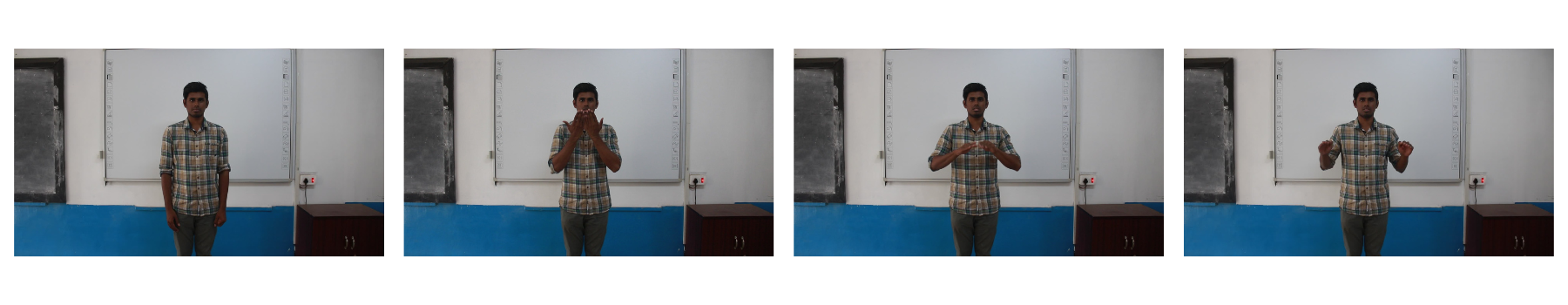}
  \caption{BdSL40 example data: Frames extracted from gesture labeled Student}
  \label{fig:figure}
\end{figure*}

\begin{figure*}
  \centering
  \includegraphics[width=\linewidth]{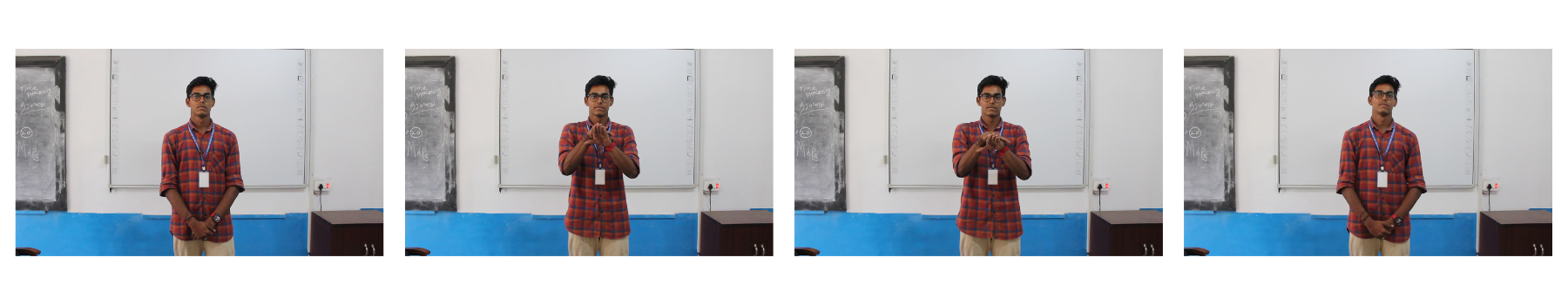}
  \caption{BdSL40 example data: Frames extracted from gesture labeled Tortoise}
  \label{fig:figure3}
\end{figure*}

\begin{figure*}
  \centering
  \includegraphics[width=\linewidth]{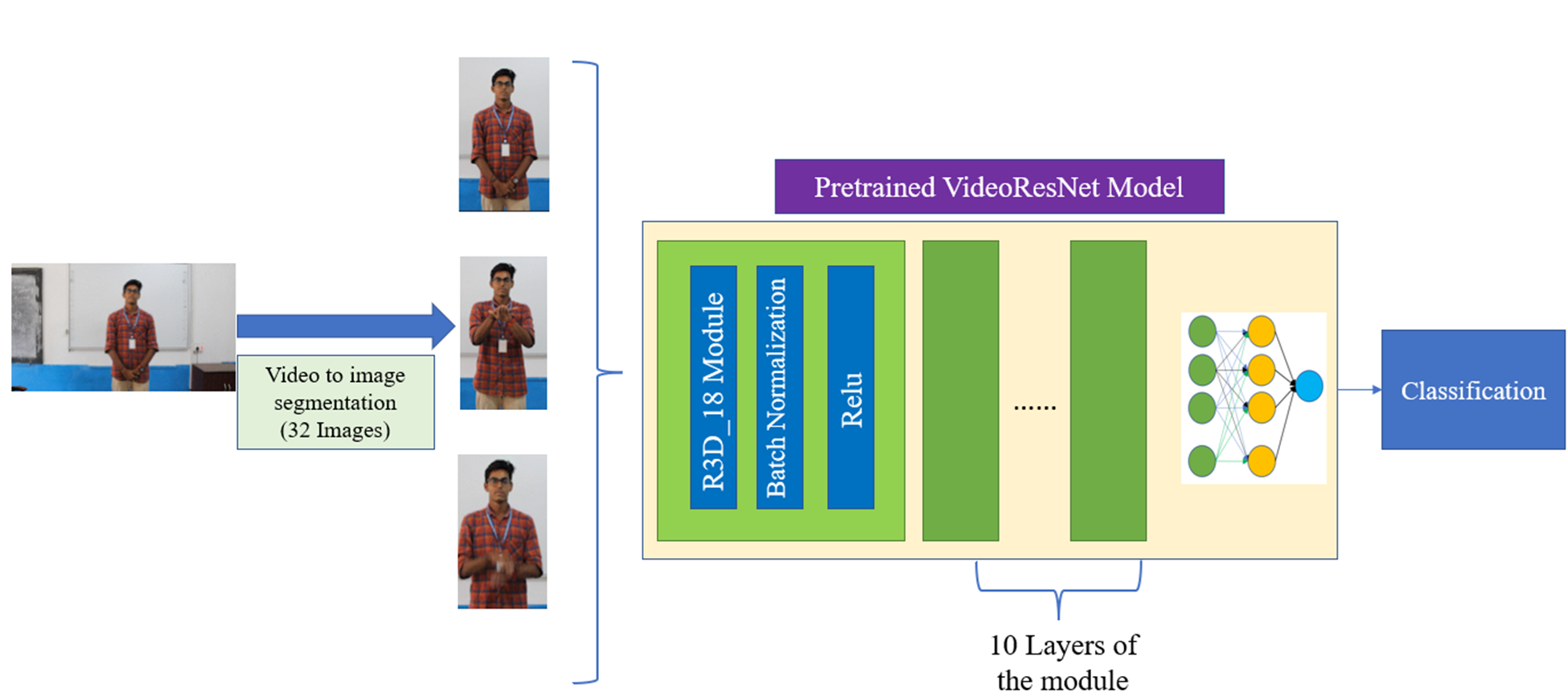}
  \caption{Classification pipeline of a specific sign language gesture. First, the frames are extracted from the video. Then they are fed into a pretrained VideoResNet Model which does the classification using 3D Convolutional Networks}
  \label{fig:figure4}
\end{figure*}

\subsection{Spatio Temporal Graph Neural Network}

In this section, we describe our methodology for classifying the Bangla Sign Language 40 (BdSL40) dataset using Spatio-Temporal Graph Neural Network.

\paragraph{Dataset Preprocessing}

For the preprocessing phase, we need to extract the hands keypoints data using a pre-trained, ready-to-use solution. Most of the available methods rely on the use of CNNs and are trained on large amounts of data. Due to its fast inference speed, high accuracy, and simple use, we chose the framework MediaPipe Hands \cite{lugaresi2019mediapipe} for our task. It extracts 21 x-y-z hand key points of the human body from each frame. The hand key point data was acquired via the Python API of MediaPipe Hands, using a minimum detection confidence of 0.8 and a minimum tracking confidence of 0.5 for tracking key points in consecutive frames.

\paragraph{Spatio-Temporal Graph Construction}

Our proposed Spatio-Temporal Graph Neural Network is fed by a four-dimensional matrix in the shape of [$N$, $C$, $T$, $V$] where $N$ denotes the batch size, $C$ denotes the number of input features(x-y-z-coordinate), $T$ denotes the input time steps and $V$ denotes the graph vertices (joints).

Our method uses the 2-stream Adaptive Graph Convolutional Network (AGCN) proposed by \citet{li2018adaptive}. The adjacency matrix for AGCN is composed of three sub-matrices: 
\begin{enumerate}
\item The inward-links starting from the wrist joint,
\item The outward links pointing in the opposite direction, and
\item The self-links of each joint.
\end{enumerate}

Thus, the matrix is of the shape [$V$, $V$, 3], where $V$ is 21 in this work. The Hand-AGCN model used in this work is a stack of 7 AGCN blocks with increasing output feature dimensions. A preceding batch normalization layer is added for input data normalization. A global average pooling layer (GAP) followed by a fully connected layer (FC) maps the output features to the corresponding output classes.

\paragraph{Hand Graph Modeling}

The MediaPipe Hands key point extraction method predicts the x-y-z coordinates of 21 hand joints; four joints per finger plus an additional wrist joint. For the definition of the underlying graph, each of the joints is connected to its natural neighbor, resulting in a graph of 21 vertices and 20 edges. These might be too few connections for the fine-grained hand movements. To obtain more semantic information in the hand graph, two types of additional joints were added. The first type of added joint links the fingertips to the base of the right neighbor finger. The second type of additional joint links the fingertips to the middle segment of the same finger. These supplementary links help to retrieve more information about the different states of the hand. The first type contains data about the horizontal and vertical distance of two fingers and can therefore help to encode the overlapping or spreading of two fingers. The second type encodes the bending of the fingers.

\begin{figure}
  \centering
  \includegraphics[width=\linewidth]{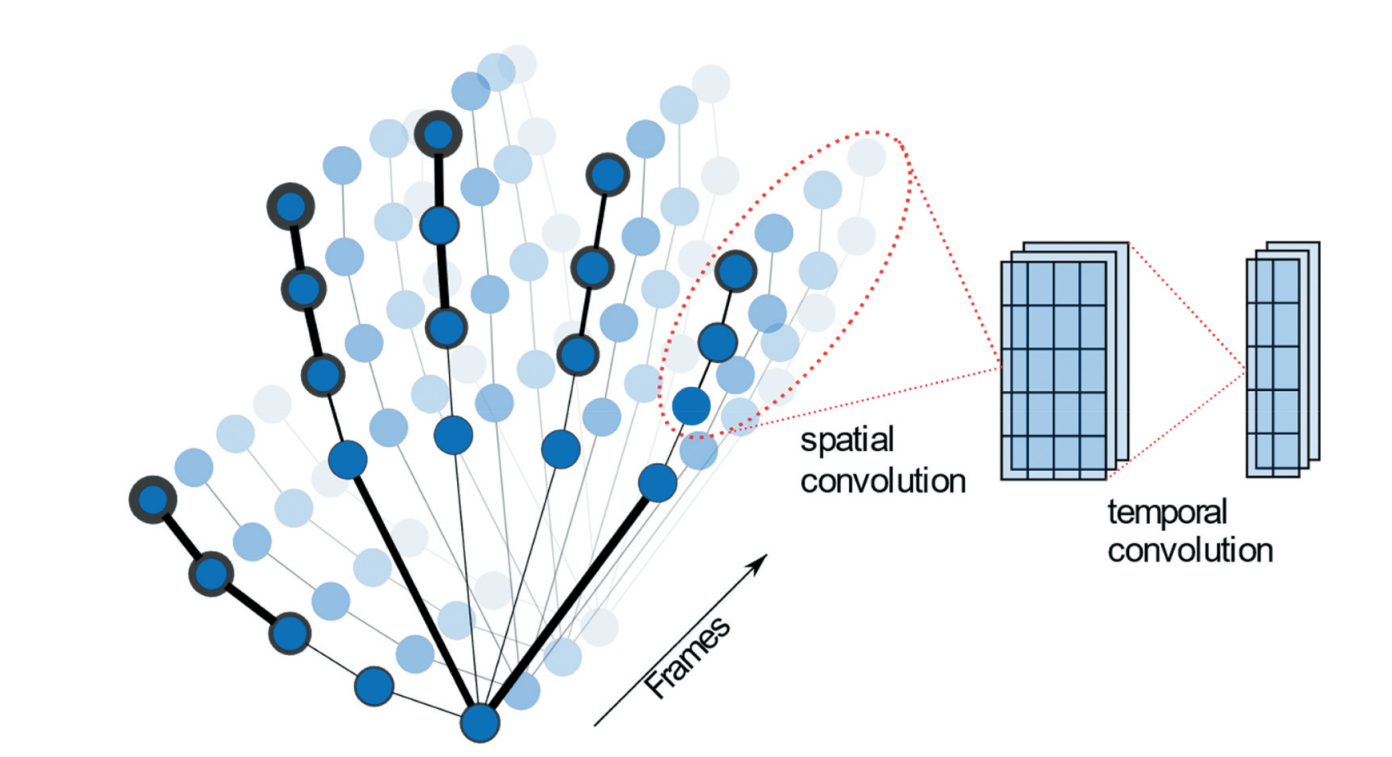}
  \caption{Spatio Temporal Graph Construction}
  \label{fig:figure2}
\end{figure}

\paragraph{Model Training}

Once the spatio-temporal graphs were constructed, they were used to train a Spatio-Temporal Graph Neural Network (GNN) model for classification. The 2s-AGCN algorithm was used for classification. 

The batch size was determined in a preliminary experiment using mini-batches of 32, 64, and 128, with 64 resulting in the highest accuracy in the validation set. The number of time steps $T$ was empirically specified to be 50 and $V$ was set to be 21. The model was trained for 5 epochs with a batch size of 64 and a learning rate of 1e-2. During the experiments, it was observed that after 5 epochs none of the models showed any substantial accuracy increase.

\section{Result Analysis}

In this study, we evaluated two methods for classifying the BdSL40 dataset: Video ResNet and Spatio-Temporal GNN. The Video ResNet model achieved an accuracy of 82.43\% On the other hand, the Spatio Temporal GNN model achieved an accuracy of 89\%

To further analyze the performance of the two methods, we also calculated precision, recall, and F1 score for each class. The results showed that the Spatio-Temporal GNN method had higher precision, recall, and F1 scores for most of the classes compared to the Video ResNet method. This suggests that the Spatio-Temporal GNN method is more effective in distinguishing between different sign language gestures.

\begin{table}
\centering
\caption{Classification results for the BdSL40 dataset using 2s-AGCN}
\label{table:results}
\begin{tabular}{lccc}
\hline
\textbf{Sign} & \textbf{Precision} & \textbf{Recall} & \textbf{F1-Score} \\ \hline
bad & 0.922 & 0.969 & 0.945 \\ \hline
book & 0.728 & 0.780 & 0.753 \\ \hline
brown & 0.887 & 0.913 & 0.900 \\ \hline
bed & 0.925 & 0.925 & 0.925 \\ \hline
camera & 0.933 & 0.944 & 0.939 \\ \hline
cheap & 0.949 & 0.944 & 0.947 \\ \hline
cow & 0.933 & 0.863 & 0.897 \\ \hline
crane & 0.446 & 0.944 & 0.606 \\ \hline
deaf & 0.915 & 0.872 & 0.893 \\ \hline
friend & 0.895 & 0.813 & 0.852 \\ \hline
fulfill & 0.906 & 0.868 & 0.886 \\ \hline
glad & 0.906 & 0.853 & 0.879 \\ \hline
heavy & 0.856 & 0.834 & 0.845 \\ \hline
i & 0.930 & 0.901 & 0.915 \\ \hline
india & 0.948 & 0.953 & 0.950 \\ \hline
lawyer & 0.989 & 0.924 & 0.956 \\ \hline
life & 0.903 & 0.914 & 0.909 \\ \hline
money & 0.915 & 0.941 & 0.928 \\ \hline
more & 0.937 & 0.886 & 0.911 \\ \hline
new & 0.913 & 0.952 & 0.932 \\ \hline
noon & 0.890 & 0.920 & 0.905 \\\hline
pant & 0.748 & 0.825 & 0.785 \\ \hline
quiet & 0.902 & 0.934 & 0.918 \\ \hline
rich & 0.898 & 0.840 & 0.868 \\ \hline
ring & 0.922 & 0.784 & 0.847 \\ \hline
shirt & 0.929 & 0.926 & 0.927 \\ \hline
shoes & 0.809 & 0.878 & 0.842 \\ \hline
skirt & 0.712 & 0.407 & 0.517 \\ \hline
soap & 0.782 & 0.539 & 0.638 \\ \hline
square & 0.926 & 0.948 & 0.937 \\ \hline
straight & 0.780 & 0.656 & 0.713 \\ \hline
student & 0.934 & 0.881 & 0.906 \\ \hline
teacher & 0.950 & 0.833 & 0.887 \\ \hline
telephone & 0.884 & 0.884 & 0.884 \\ \hline
thick & 0.955 & 0.923 & 0.939 \\ \hline
time & 0.605 & 0.721 & 0.658 \\ \hline
tortoise & 0.822 & 0.694 & 0.753 \\ \hline
winter & 0.891 & 0.936 & 0.913 \\ \hline
yesterday & 0.863 & 0.852 & 0.857 \\ \hline
you & 0.856   &  0.633  &  0.728 \\ \hline
\end{tabular}
\end{table}

\newpage
\bibliography{refs}

\begin{thebibliography}{7}
\providecommand{\natexlab}[1]{#1}

\bibitem[{Johnson and Johnson(2016)}]{johnson2016distinction}
Johnson, R.~J.; and Johnson, J.~E. 2016.
\newblock Distinction between west Bengal sign language and Indian sign
  language based on statistical assessment.
\newblock \emph{Sign Language Studies}, 16(4): 473--499.

\bibitem[{Li et~al.(2018)Li, Wang, Zhu, and Huang}]{li2018adaptive}
Li, R.; Wang, S.; Zhu, F.; and Huang, J. 2018.
\newblock Adaptive graph convolutional neural networks.
\newblock In \emph{Proceedings of the AAAI conference on artificial
  intelligence}, volume~32.

\bibitem[{Lugaresi et~al.(2019)Lugaresi, Tang, Nash, McClanahan, Uboweja, Hays,
  Zhang, Chang, Yong, Lee et~al.}]{lugaresi2019mediapipe}
Lugaresi, C.; Tang, J.; Nash, H.; McClanahan, C.; Uboweja, E.; Hays, M.; Zhang,
  F.; Chang, C.-L.; Yong, M.~G.; Lee, J.; et~al. 2019.
\newblock Mediapipe: A framework for building perception pipelines.
\newblock \emph{arXiv preprint arXiv:1906.08172}.

\bibitem[{Sridhar et~al.(2020)Sridhar, Ganesan, Kumar, and
  Khapra}]{sridhar2020include}
Sridhar, A.; Ganesan, R.~G.; Kumar, P.; and Khapra, M. 2020.
\newblock Include: A large scale dataset for indian sign language recognition.
\newblock In \emph{Proceedings of the 28th ACM international conference on
  multimedia}, 1366--1375.

\bibitem[{Tran et~al.(2015)Tran, Bourdev, Fergus, Torresani, and
  Paluri}]{tran2015learning}
Tran, D.; Bourdev, L.; Fergus, R.; Torresani, L.; and Paluri, M. 2015.
\newblock Learning spatiotemporal features with 3d convolutional networks.
\newblock In \emph{Proceedings of the IEEE international conference on computer
  vision}, 4489--4497.

\bibitem[{Tran et~al.(2018)Tran, Wang, Torresani, Ray, LeCun, and
  Paluri}]{tran2018closer}
Tran, D.; Wang, H.; Torresani, L.; Ray, J.; LeCun, Y.; and Paluri, M. 2018.
\newblock A closer look at spatiotemporal convolutions for action recognition.
\newblock In \emph{Proceedings of the IEEE conference on Computer Vision and
  Pattern Recognition}, 6450--6459.

\bibitem[{Yu, Yin, and Zhu(2017)}]{yu2017spatio}
Yu, B.; Yin, H.; and Zhu, Z. 2017.
\newblock Spatio-temporal graph convolutional networks: A deep learning
  framework for traffic forecasting.
\newblock \emph{arXiv preprint arXiv:1709.04875}.

\end{thebibliography}

\end{document}